%% file: root.tex
\documentclass[letterpaper, 10 pt, conference]{ieeeconf}  

\IEEEoverridecommandlockouts                              

\usepackage[utf8]{inputenc}
\usepackage[english]{babel}

\usepackage{xcolor}
\usepackage{graphicx}
\graphicspath{ {figures/} }
\usepackage{booktabs} 
\usepackage{amsmath}
\usepackage{multirow}
\usepackage{authblk}
\usepackage{hyperref}

\usepackage{subfiles} 

\title{\LARGE \bf Learning to Infer Kinematic Hierarchies for Novel Object Instances}

\makeatletter \renewcommand\AB@affilsepx{\quad \protect\Affilfont} \makeatother

\author[2]{Hameed Abdul-Rashid}
\author[1]{Miles Freeman}
\author[1]{Ben Abbatematteo}
\author[1]{George Konidaris}
\author[1]{Daniel Ritchie}
\affil[1]{\small{Brown University}}
\affil[2]{\small{University of Illinois at Urbana-Champaign}}

\newcommand{\etal}{\textit{et al.} }
\newcommand{\etalN}{\textit{et al.}}

\begin{document}

\maketitle

\thispagestyle{empty}
\pagestyle{empty}

\begin{abstract}
\input{sections/00-abstract}
\end{abstract}

\input{sections/01-intro}
\input{sections/02-related}
\input{sections/03-approach}
\input{sections/04-method}

\input{sections/05-results}
\input{sections/06-future}
\input{sections/07-conclusion}

\bibliography{root}
\bibliographystyle{plain}
\end{document}

%% file: sections/00-abstract.tex
Manipulating an articulated object requires perceiving its \emph{kinematic hierarchy}: its parts, how each can move, and how those motions are coupled.
Previous work has explored perception for kinematics, but none infers a complete kinematic hierarchy on never-before-seen object instances, without relying on a schema or template.
We present a novel perception system that achieves this goal.
Our system infers the moving parts of an object and the kinematic couplings that relate them.
To infer parts, it uses a point cloud instance segmentation neural network and 
to infer kinematic hierarchies, it uses a graph neural network to predict the existence, direction, and type of edges (i.e. joints) that relate the inferred parts.
We train these networks using simulated scans of synthetic 3D models.
We evaluate our system on simulated scans of 3D objects, and we demonstrate a proof-of-concept use of our system to drive real-world robotic manipulation.

%% file: sections/01-intro.tex


\section{Introduction}
\label{sec:intro}

People frequently interact with articulated objects: opening doors, putting things away in drawers, adjusting the height of chairs, etc.
To a human, such interactions are effortless, but the underlying process is complex: one must perceive the \emph{kinematic hierarchy} of the object (i.e. which parts can move, how they move, and how those motions are coupled). Then, one must plan a sequence of actions to transform the object from its current kinematic pose into another, and finally, manipulate the object to execute that plan.
The perception part of this process is particularly challenging, as objects with similar function can vary considerably in both their structure and their geometry. Cabinets, for example, can have differing numbers of drawers in different arrangements and many different doorknob shapes.
To succeed at manipulating such objects in the wild, an agent must infer the kinematic hierarchies of never-before-seen object instances with these kinds of variations.

This problem has received considerable attention in computer vision and robotics literatures.
Vision researchers have made progress on detecting atomic parts of 3D objects and predicting the presence and parameters of individual kinematic joints~\cite{Shape2Motion,DeepPart,RPMNet}.
These approaches are trained on large datasets and thus have the potential to generalize to novel objects from a known category,
however, they do not infer kinematic hierarchies.
Roboticists have leveraged embodiment to solve the problem by providing the agent with human demonstrations or allowing it to interact with its environment~\cite{sturm2011probabilistic, katz2011interactive, kroemer2019review}.
Recent works infer kinematic hierarchies but assume a simple template (e.g. one door per cabinet) and thus cannot generalize to the structural variability that exists in real world objects ~\cite{abbatematteo2019learning,li2019category,jain2020screwnet}.

\begin{figure}[t!]
    \centering
    \includegraphics[width=\linewidth]{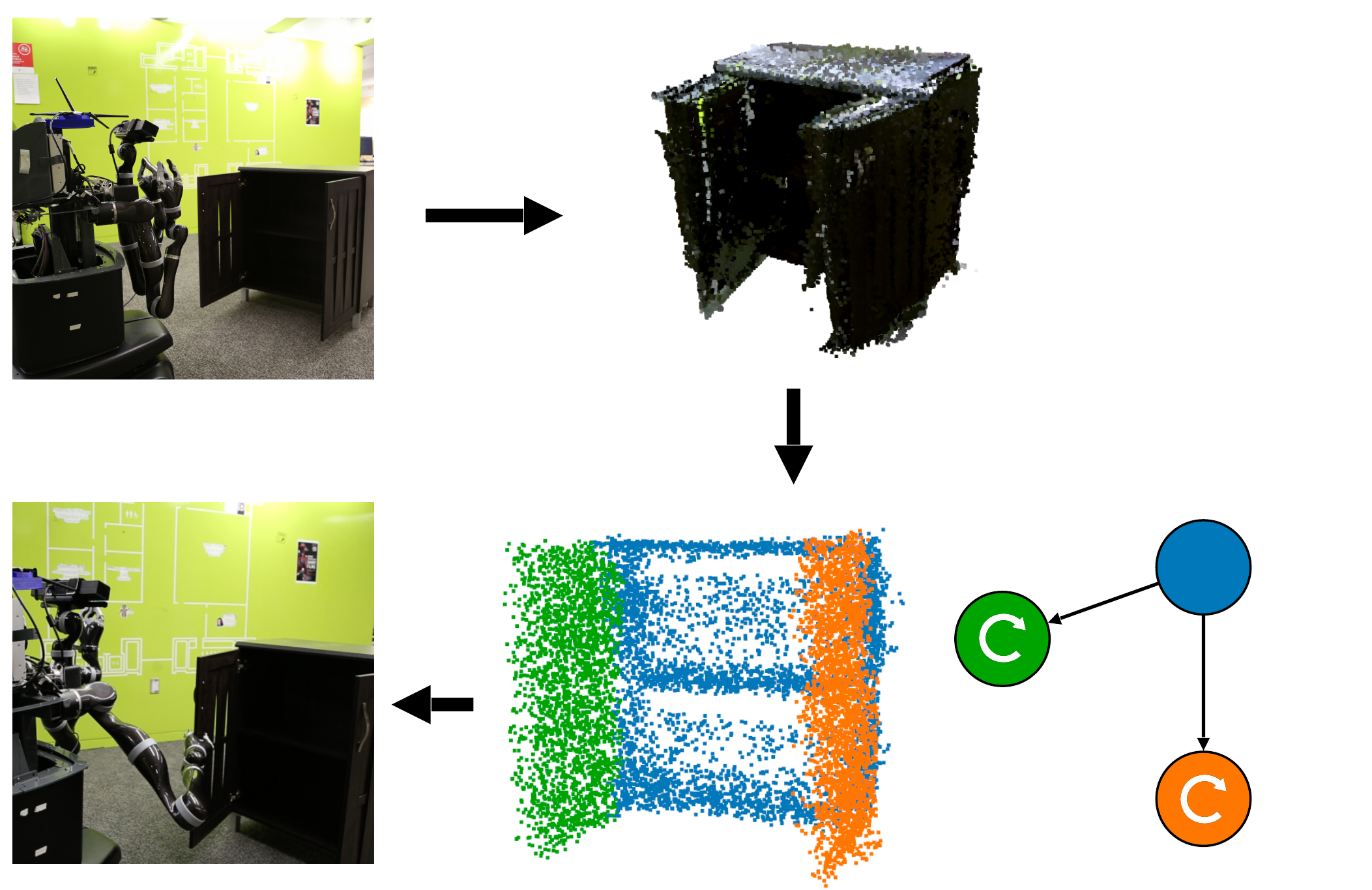}
    \caption{
    We propose a system for robot perception of the kinematic structure of never-before-seen object instances.
    The robot scans the object from multiple viewpoints, building a 3D point cloud representation of it.
    Our system then segments this point cloud into parts and infers the kinematic hierarchy that couples them, allowing for the robot to plan and execute articulated motions of the object.
    }
    \label{fig:teaser}
\end{figure}

In this paper, we propose a perception system that both infers kinematic hierarchies and also generalizes to novel object geometries and structures.
Our system infers the moving parts of an object and the kinematic couplings between them.
To infer parts, it uses a point cloud instance segmentation neural network.
To infer kinematic hierarchies, it uses a graph neural network which learns to infer the existence, direction, and type of edges (i.e. joints) that relate the inferred parts.
This system is trained on simulated scans of a large set of synthetic 3D models, enabling it to learn to generalize to new, never-before-seen object instances which may differ geometrically and structurally from the training data.
We evaluate our system's perception abilities on synthetic point clouds, and find that our system reliably detects moving parts and reconstructs kinematic hierarchies on synthetic data.
We also demonstrate a proof-of-concept application of using the output of our perception system to drive manipulation executed by a real-world robot platform.


\begin{figure*}[t!]
    \includegraphics[width=\textwidth]{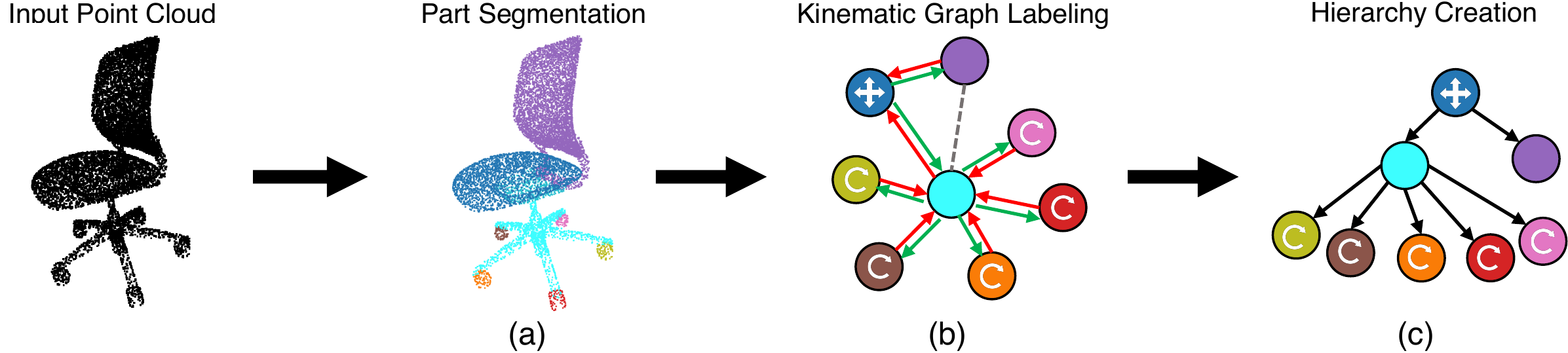}
    \caption{
    Overview of our system.
    Given an input 3D point cloud representation of an object, our system (a) segments the point cloud into parts using an instance segmentation network with multi-view consensus, (b) constructs a part graph and uses a graph neural network to label its nodes and edges with kinematic properties, and (c) constructs a kinematic hierarchy from this graph which can be used for robot manipulation.
    }
    \label{fig:approach}
\end{figure*}

In summary, our contributions are:
\begin{itemize}
    \item The first method for inferring kinematic hierarchies from never-before-seen instances of 3D objects, without reliance on a schema or template.
    \item A novel graph neural network approach for predicting hierarchies from part segments.
\end{itemize}




%% file: sections/02-related.tex


\section{Related Work}
\label{sec:related}


\subsection{Approaches from the Robotics Community}
Kinematic model and articulated pose estimation are well-studied problems in the robotics literature. Observation-based approaches~\cite{sturm2011probabilistic, pillai2015learning, niekum2015online, Daniele2017LearningAO, yan2006automatic} fit motion models to an observed sequence of part poses. Interactive approaches leverage the embodiment of the manipulator to generate part motion autonomously~\cite{katz2011interactive, hausman2015active, barragan2014interactive, martinbrockinteract, kulick2015active, lockboxphysical} and to reason about the intended effects of actions \cite{jain2019learning}. More recently, learning-based approaches to category-level model and pose estimation~\cite{li2019category, abbatematteo2019learning, jain2020screwnet, FormNet} have shown promise, but are thus-far limited to a-priori known connectivity graphs. Several methods estimate only articulated pose given known kinematic models and geometry~\cite{brookshire2016articulated, Desingheaaw4523, pavlasek2020parts}. 

\subsection{Approaches from the Computer Vision Community}
Wang \etal train a model to jointly predict movable parts and their motion parameters from an input point cloud.
\cite{Shape2Motion}.
Yi \etal take a pair of articulated 3D shapes of the same category but in different kinematic poses, and jointly infer moving parts and their kinematic motion parameters~\cite{DeepPart}.
Yan \etal infer movable parts by hallucinating their motions over multiple time steps~\cite{RPMNet}.
These approaches train deep neural networks which generalize to geometrically and structurally different input objects.
However, none infer the structure of the object's kinematic hierarchy: they treat each articulated part as it if moves in isolation.
Our approach bears resemblance to RigNet, a system for inferring the skeletal joint structure of articulated humanoid and creature characters~\cite{RigNet}.
Like us, they also use a graph neural network for predicting graph topology information.


%% file: sections/03-approach.tex


\section{Approach}
\label{sec:approach}

In this section, we overview our method for kinematic mobility perception.
Fig.~\ref{fig:approach} shows a schematic of our pipeline.

\subsection{Part Segmentation}     
The input to our system is an unlabeled point cloud depicting an articulated object; a robot can obtain such input by unprojecting and consolidating frames from an onboard depth sensor.
Our system first segments this point cloud into parts, some of which are movable (Fig.~\ref{fig:approach}a).
This stage relies upon an existing neural network architecture for point cloud instance segmentation (i.e. segmenting a point cloud without consistent part labels)~\cite{PartNet}.


\subsection{Kinematic Graph Labeling} 
Given a part-segmented point cloud, our system next infers the articulation properties of those parts: the type of articulated motion they support (if any) and how the motions of different parts are coupled.
To do this, it converts the segmented point cloud into a graph and phrases the problem as one of \emph{graph labeling}: nodes in the graph represent parts, and edges between them represent potential kinematic connections (e.g. joints) between parts.
A graph neural network predicts whether each node is static, rotating, or translating, as well as whether each edge should exist (i.e. whether two parts should be kinematically connected), the probability of each node being the root of the tree, and the direction of kinematic dependency (Fig.~\ref{fig:approach}b).
                                                             
\subsection{Hierarchy Creation}
Given the labeled graph, we construct a kinematic hierarchy by converting the bidirectional graph into a n-ary tree. Xu \etal address this problem by constructing a pairwise matrix where each entry represents the negative log probability of the corresponding parts possessing an edge connection~\cite{RigNet}.  The pairwise matrix is traversed from a predicted root node and a n-ary tree is extracted via a Minimum Spanning Tree (MST) algorithm ~\cite{MST}.  We adopt this method to convert our predicted bidirectional graphs to an n-ary tree.(Fig.~\ref{fig:approach}c).


%% file: sections/04-method.tex


\section{Part Segmentation}
\label{sec:partseg}


To segment point clouds into parts, we use a neural network based on that of PartNet~\cite{PartNet}.
This architecture uses a PointNet++ network~\cite{PointNet++} to compute per-point embeddings, each of which is then fed to a semantic segmentation branch (assigns semantic labels to each point) and an instance segmentation branch (assign each point to one of $N$ possible instance masks; we use $N=24$).
As our system does not assume parts are semantically labeled, we could remove the semantic segmentation branch.
However, PartNet showed that semantic segmentation helps improve the quality of instance segmentation.
Thus, we treat the semantic segmentation branch as a ``motion type prediction'' branch (static, rotating, translating, rotating and translating)---labels which we do assume of our data.
We train the segmentation network on 3D objects from the PartNet-Mobility dataset~\cite{SAPIEN}.
For augmentation, we use multiple poses for each moving part of each object, as described in more detail in Section~\ref{sec:results}.
To simulate a robot's onboard depth sensor, we render point clouds from these objects using a software simulator of the Kinect sensor~\cite{KinectSimulator}.
We train the network using the Adam optimizer~\cite{Adam} with learning rate $10^{-4}$.
Training was run for at most 8 hours per category.

\begin{figure}[t!]
    \centering
    \setlength{\tabcolsep}{6pt}
    \renewcommand{\arraystretch}{3.5}
    \begin{tabular}{ccc}
         Part Segments & Initial Graph & GT Tree
         \\
         \includegraphics[width=0.3\linewidth]{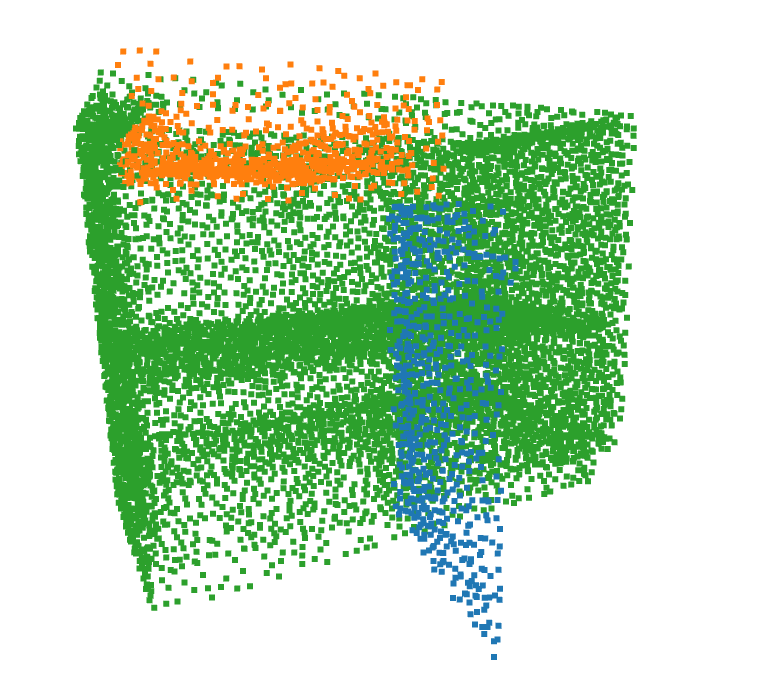} &
         \includegraphics[width=0.25\linewidth]{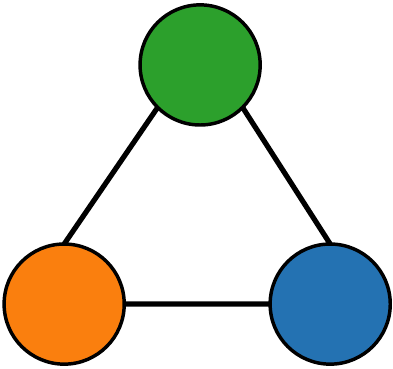} &
         \includegraphics[width=0.25\linewidth]{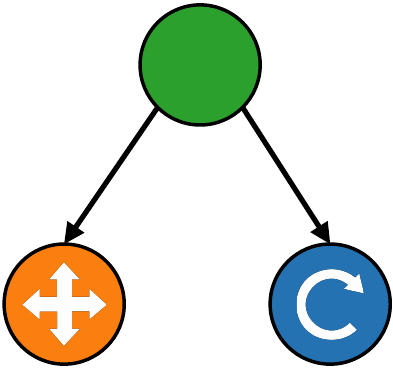}
         \\
         \includegraphics[width=0.3\linewidth]{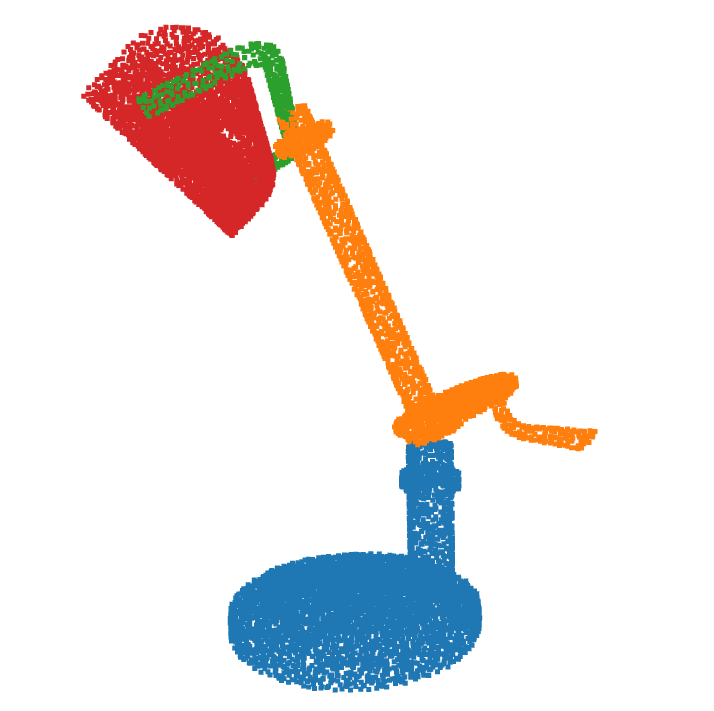} &
         \includegraphics[width=0.25\linewidth]{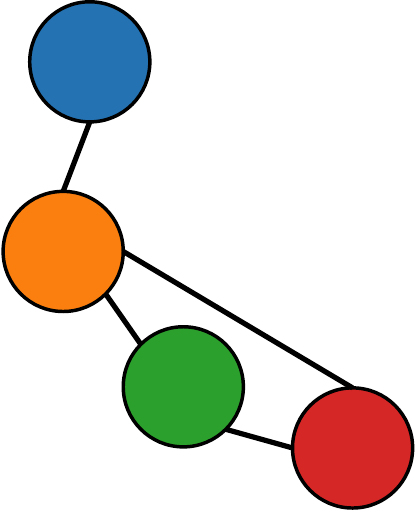} &
         \includegraphics[width=0.25\linewidth]{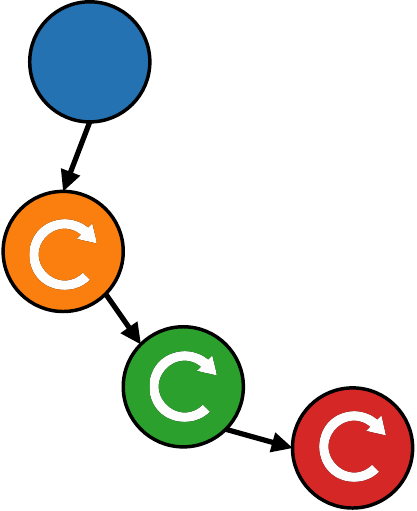}
         \\
         \includegraphics[width=0.3\linewidth]{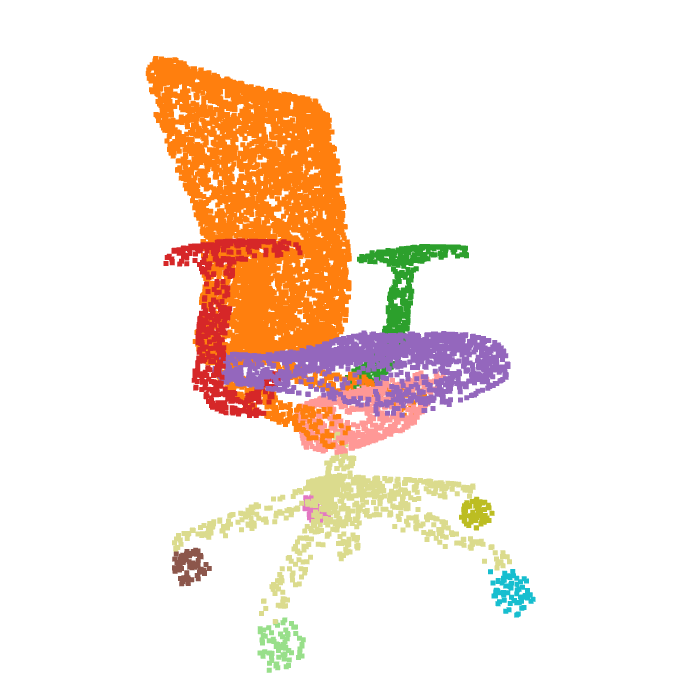} &
         \includegraphics[width=0.28\linewidth]{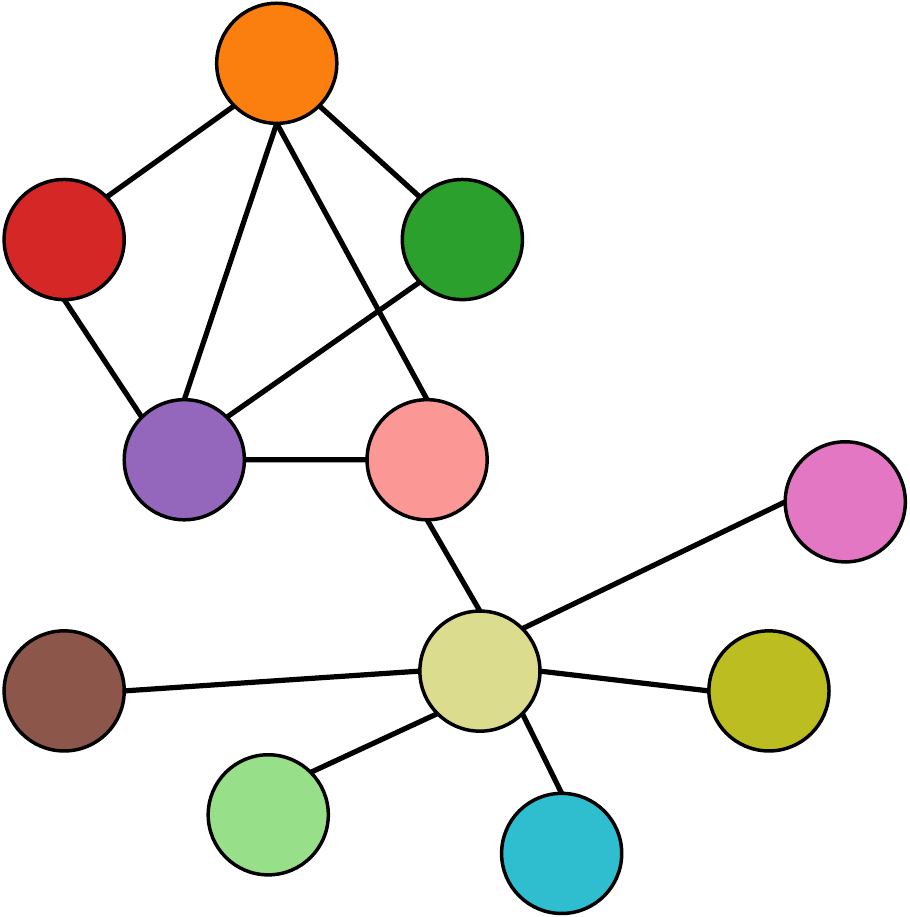} &
         \includegraphics[width=0.28\linewidth]{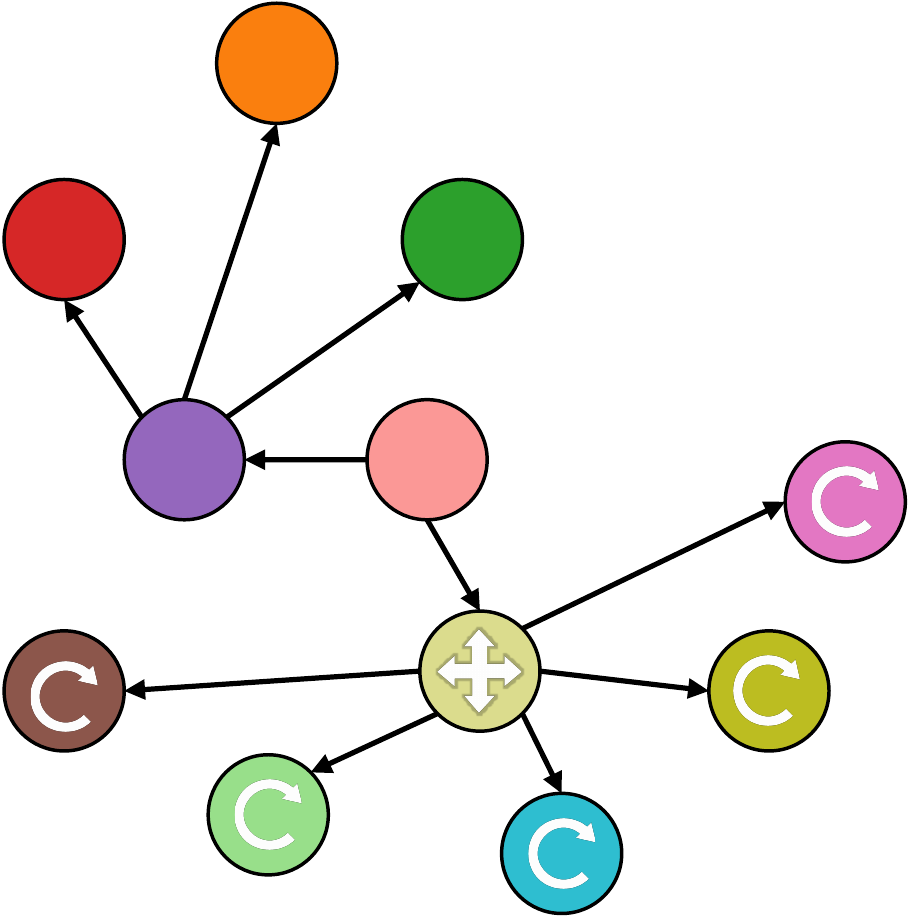}
    \end{tabular}
    \caption{
    Segmented point clouds, their initial overcomplete part graphs, and their ground truth part graphs (as determined by a human labeler).
    }
    \label{fig:graph_construction}
\end{figure}

\begin{figure*}[t!]
    \centering
    \includegraphics[width=\linewidth]{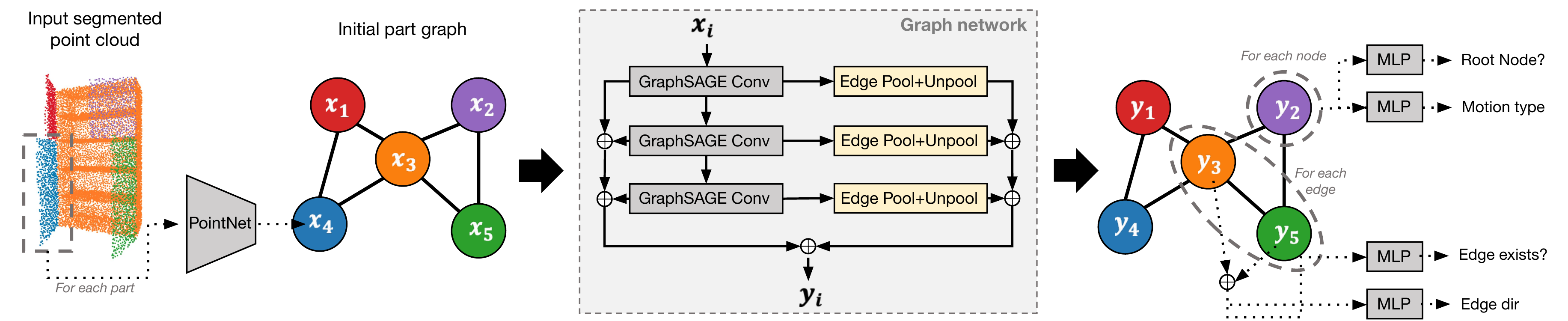}
    \caption{
    Neural network architecture for kinematic graph labeling.
    A PointNet~\cite{PointNet_Qi} converts each part point cloud into an initial node feature $\mathbf{x}$.
    These features go through graph convolutional + pooling layers to produce features $\mathbf{y}$ which are passed to MLPs for predicting node and edge attributes.
    }
    \label{fig:gnn_arch}
\end{figure*}

\section{Kinematic Graph Annotation}
\label{sec:graphannotate}

Here we describe how our system takes a segmented point cloud, converts it to a graph, and labels this graph with the information required to construct a kinematic hierarchy.

\subsection{Point Cloud to Graph Conversion}
The kinematic hierarchy for a segmented object is a directed tree: tree nodes correspond to parts, nodes are labeled with a motion type (rotation and/or translation)
and edge directions indicate kinematic couplings (i.e. edge $A \rightarrow B$ means part $B$ moves with part $A$).
To infer this tree from a segmented point cloud, our system first constructs an overcomplete, undirected, unlabeled graph over the parts, and then prunes edges, determines edge directions, and labels nodes using a graph neural network.
Our system constructs the initial graph by adding an undirected edge between any two parts whose Euclidean distance is below a small threshold. 
Fig.~\ref{fig:graph_construction} shows some example segmented point clouds and their initial graphs.

\subsection{Graph Labeling Network}
Given a graph constructed by the above procedure, we use a graph neural network to label its nodes and edges with the information required to construct a kinematic hierarchy.

\noindent
\newline
\textbf{Input features:}
Every node corresponds to one part, i.e. one subset of points from the object point cloud.
We encode these point subsets into per-node vectors $\mathbf{x}$ using a PointNet~\cite{PointNet_Qi}.

\noindent
\newline
\textbf{Graph network:}
To make per-node and per-edge predictions, our system must understand the context of each node or edge within the object.
To satisfy this goal, we use a graph convolutional network to convert node features $\mathbf{x}$ into context-aware features $\mathbf{y}$ (Fig.~\ref{fig:gnn_arch}).
Our network passes all nodes through three GraphSAGE convolution layers~\cite{GraphSAGE}, which compute new node features  based on learned aggregations of neighbor node features.
Each convolution layer is followed by a pooling and unpooling layer based on learned edge collapses~\cite{EdgePooling}.
In early experiments, we found this pooling + unpooling scheme to perform better than architectures that used only convolution.
Finally, the outputs of all convolution layers and all pooling layers are concatenated to form a multi-scale feature vector $\mathbf{y}$ for each node.


\noindent
\newline
\textbf{Node and edge attribute prediction:}
The per-node features $\mathbf{y}$ are fed to multi-layer perceptrons (MLPs) for predicting node attributes.
For edge attributes, the $\mathbf{y}$'s for both edge endpoints are concatenated and fed to the MLP.

\subsection{Training}
We train the graph labeling network on synthetic 3D models from PartNet-Mobility~\cite{SAPIEN}.
We first pre-train the node feature PointNet by training it as the encoder in an autoencoder framework; this network trains for 500 epochs using Nesterov SGD with a learning ratte of $10^{-3}$.
The graph labeling network trains for 30 epochs using Adam with a learning rate of $10^{-3}$.
Train both networks takes 1.5 - 3 hours on an NVIDIA RTX 2080 Ti.

%% file: sections/05-results.tex
\section{Results \& Evaluation}
\label{sec:results}

Here we evaluate our method's ability to infer kinematic hierarchies via controlled experiments and a demonstration on a real-world robot.
For experiments on synthetic PartNet-Mobility objects, we use the same train/test splits introduced by Mo~\etalN~\cite{PartNet}, restricted to models with valid kinematic trees. We augment the training and test datasets by uniformly sampling articulated poses within each part's ground truth range of motion. We train on 249 storage furniture models, 22 lamp models, and 53 chair models; we test on 69, 8, and 12 models, respectively. We use 18 pose augmentations for training and testing.



\subsection{Part Segmentation}
\begin{figure*}[t!]
    \centering
    \vspace{0.2cm}
    \setlength{\tabcolsep}{1pt}
    \begin{tabular}{ccccccc}
        & \multicolumn{3}{c}{\rule[1.5pt]{10em}{0.8pt} Clean \rule[1.5pt]{10em}{0.8pt}} & \multicolumn{3}{c}{\rule[1.5pt]{10em}{0.8pt} Noisy \rule[1.5pt]{10em}{0.8pt}}
        \\
        \raisebox{1em}{\rotatebox{90}{Predicted Parts}} &
        \includegraphics[width=.15\linewidth]{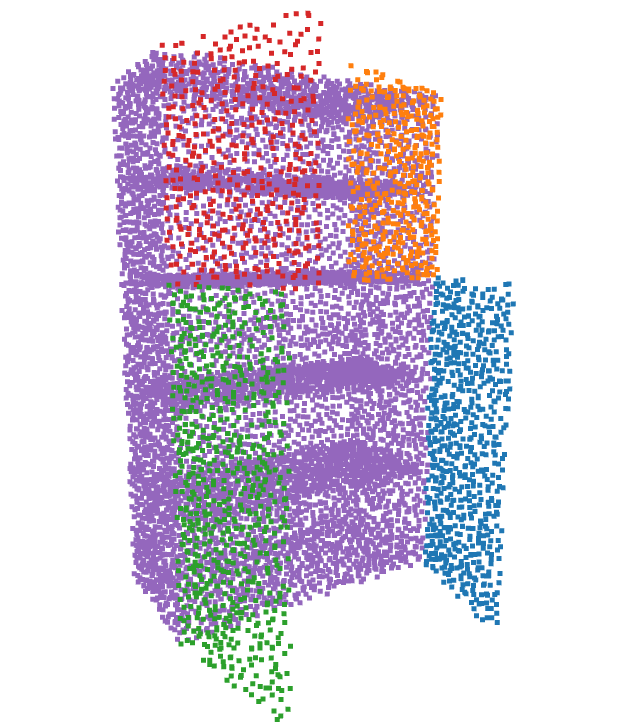} &
        \includegraphics[width=.15\linewidth]{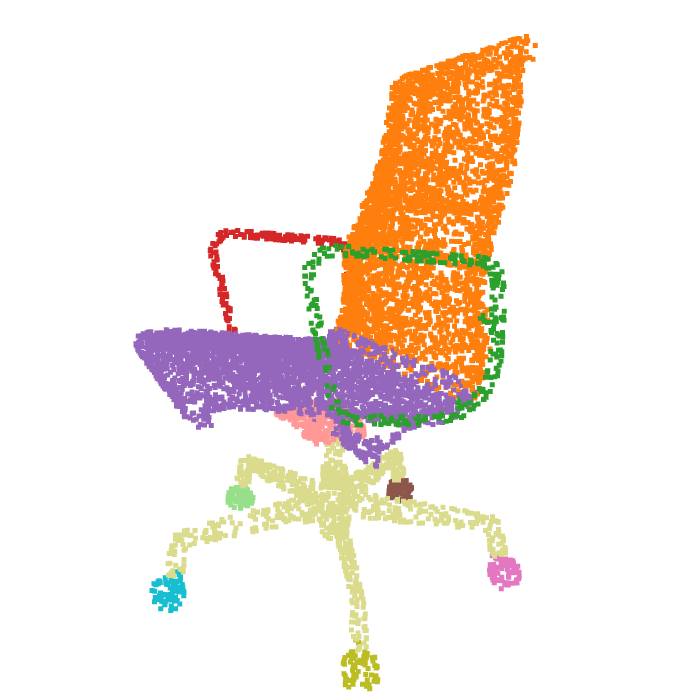} &
        \includegraphics[width=.15\linewidth]{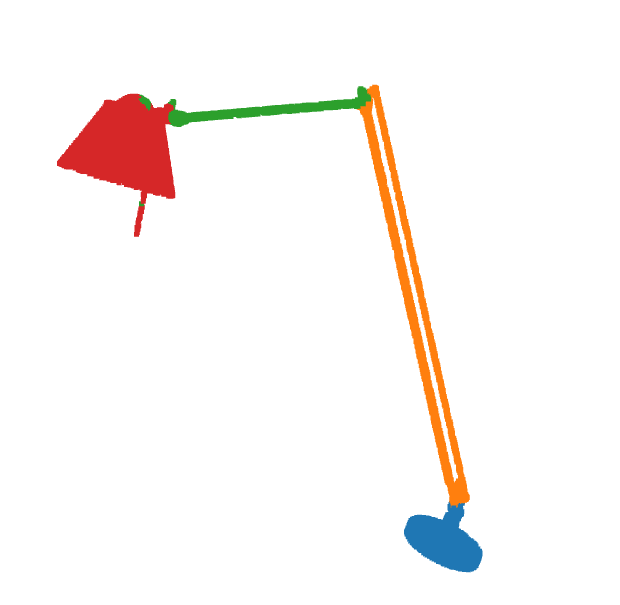} &
        \includegraphics[width=.125\linewidth]{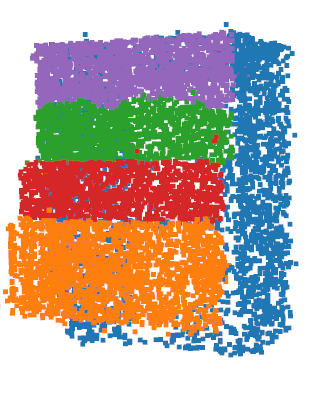} &
        \includegraphics[width=.15\linewidth]{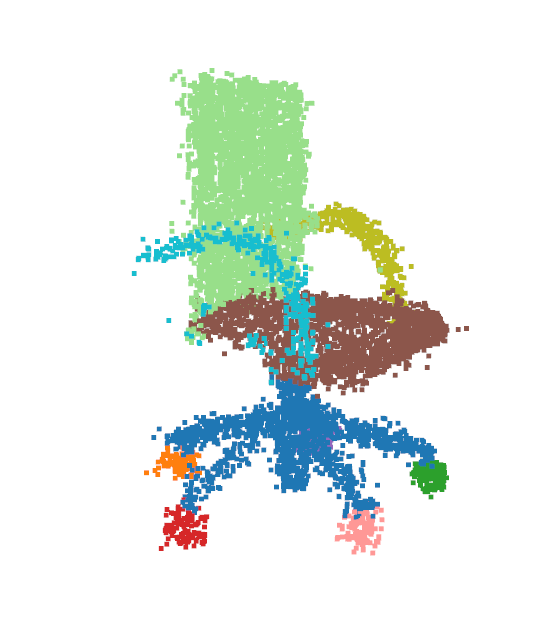} &
        \includegraphics[width=.15\linewidth]{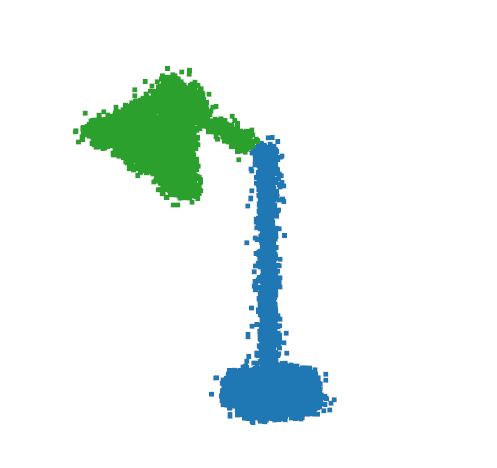}
        
        \\
        \raisebox{1.5em}{\rotatebox{90}{Target Parts}} &
        \includegraphics[width=.15\linewidth]{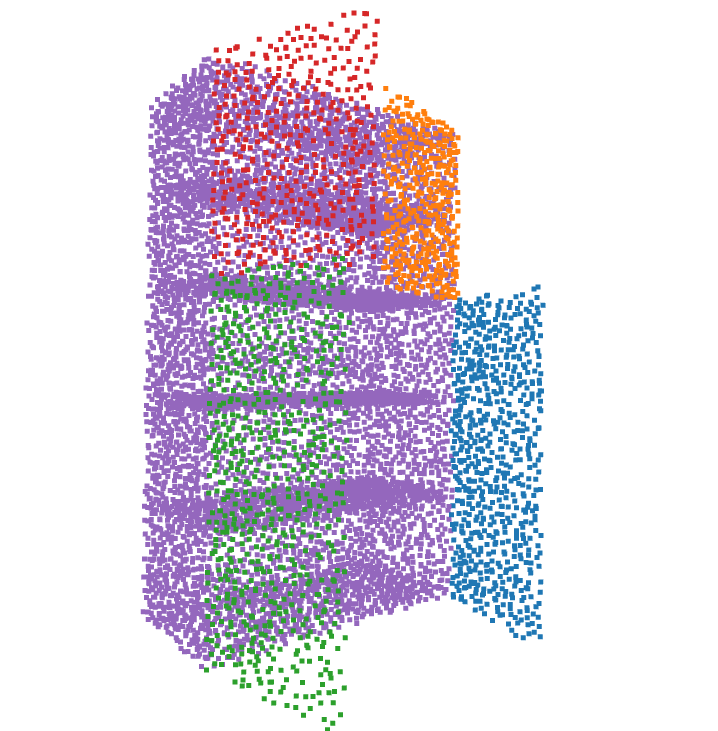} &
        \includegraphics[width=.15\linewidth]{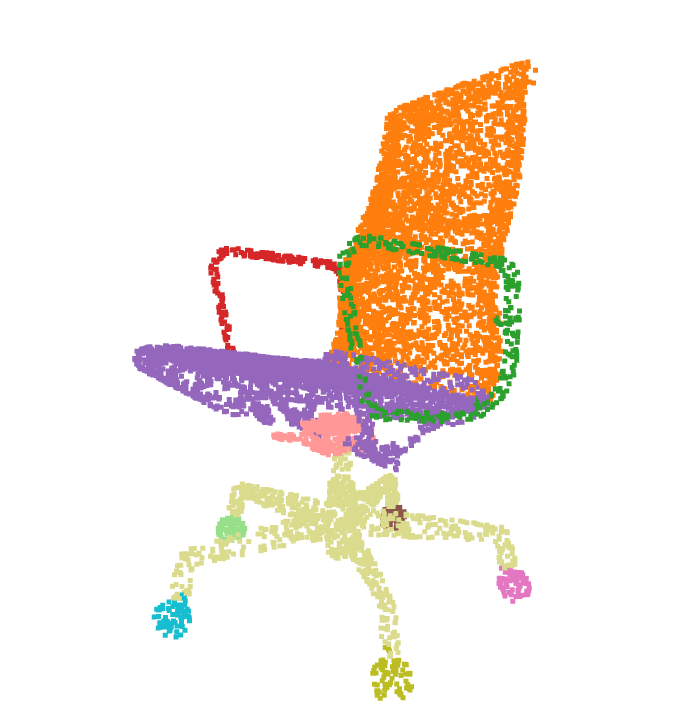} &
        \includegraphics[width=.15\linewidth]{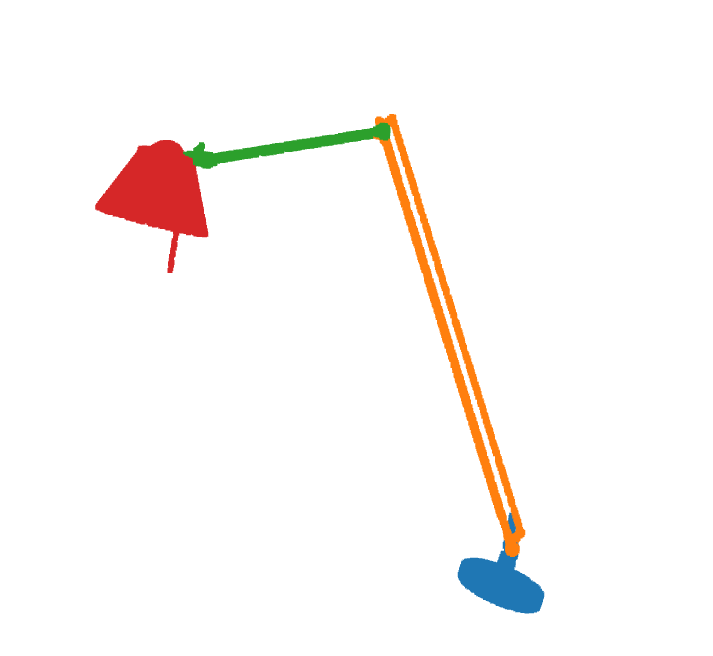} &
        \includegraphics[width=.15\linewidth]{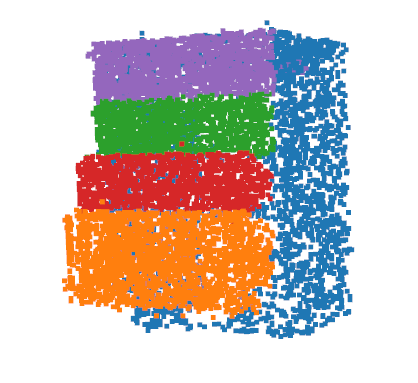} &
        \hspace{0.5cm}
        \includegraphics[width=.125\linewidth]{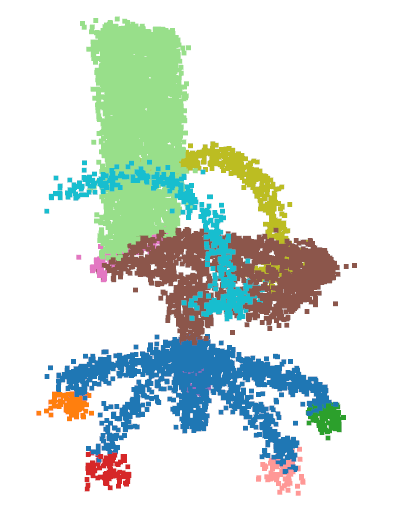} &
        \includegraphics[width=.15\linewidth]{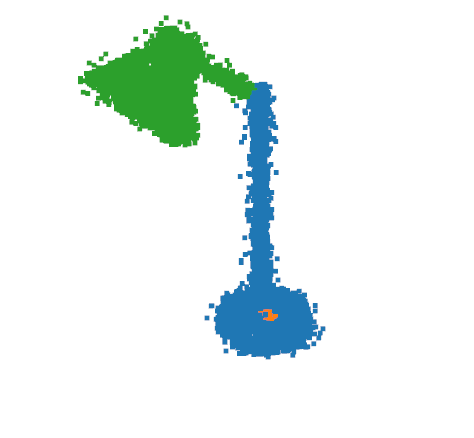}
    \end{tabular}
    \caption{
    Examples of part segmentation for both Clean and Noisy point clouds.
    }
    \label{fig:segmentation_results}
\end{figure*}

Table~\ref{tab:segmentation_results} summarizes the performance of our part segmentation method on synthetic 3D objects.
We use an established instance segmentation metric~\cite{PartNet}: we compute the intersection over union (IoU) of each predicted part segment with its closest ground truth part segment, classify it as a ``correct'' prediction if the IoU is over 0.5, and then compute the mean average precision of these classifications across all parts in all test set objects.
We compute this metric under two conditions: (1) \emph{Clean} point clouds sampled directly from the surface of PartNet-Mobility 3D meshes, (2) \emph{Noisy} point clouds generated by simulated Kinect scans of PartNet-Mobility models as described in Section~\ref{sec:partseg}.
Fig.~\ref{fig:segmentation_results} shows some qualitative examples.
The network reliably segments parts in clean point clouds, and this performance degrades slightly in the noisy depth scans. 
The lamp category is especially challenging due to the small number of models present in the training data. 


\begin{table}[t!]
    \centering
    \caption{
    Part segmentation on PartNet-Mobility (mAP @ 0.5 IoU)
    }
    \begin{tabular}{lccc}
        \toprule
        \textbf{Category} & \textbf{Clean} & \textbf{Noisy} 
        \\
        \midrule
        \emph{Storage Furniture} & 0.922 & 0.907 
        \\
        \emph{Lamp} & 0.695 & 0.593 
        \\
        \emph{Chair} & 0.824 & 0.811 
        \\
        \bottomrule
    \end{tabular}
    \label{tab:segmentation_results}
\end{table}

\subsection{Kinematic Structure Prediction}
\begin{figure*}[t!]
    \centering
    \setlength{\tabcolsep}{1pt}
    \begin{tabular}{ccccccc}
        & \multicolumn{3}{c}{\rule[1.5pt]{10em}{0.8pt} Clean \rule[1.5pt]{10em}{0.8pt}} & \multicolumn{3}{c}{\rule[1.5pt]{10em}{0.8pt} Noisy \rule[1.5pt]{10em}{0.8pt}}
        \\
        \raisebox{3em}{\rotatebox{90}{Parts}} &
        \includegraphics[width=.15\linewidth]{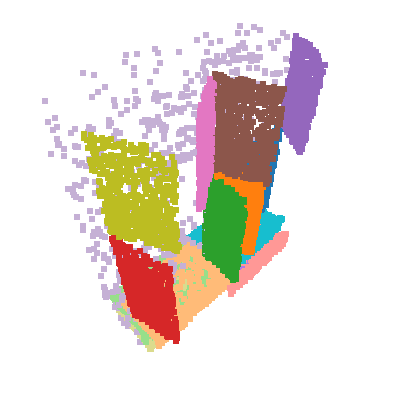} &
        \includegraphics[width=.15\linewidth]{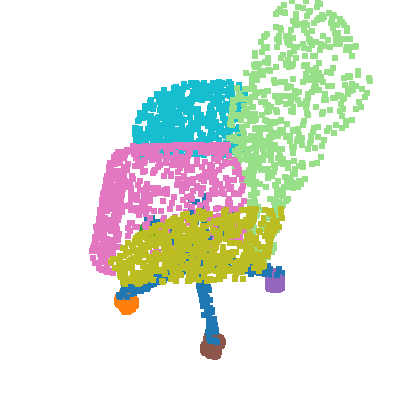} &
        \includegraphics[width=.15\linewidth]{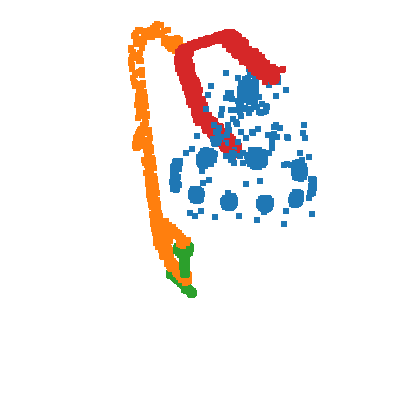} &
        \includegraphics[width=.15\linewidth]{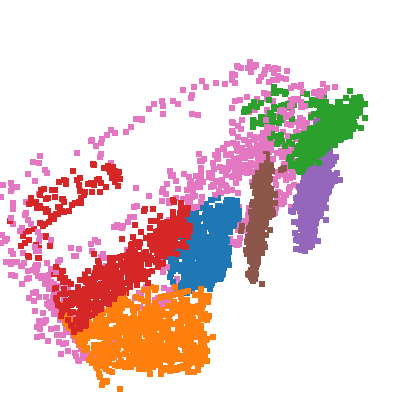} &
        \includegraphics[width=.15\linewidth]{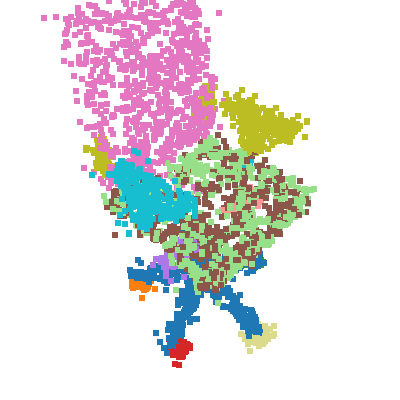} &
        \includegraphics[width=.15\linewidth]{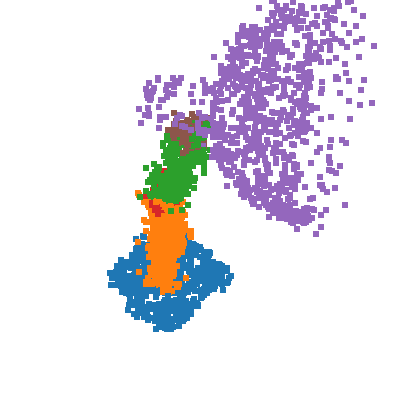} 
        \\
        \raisebox{1em}{\rotatebox{90}{Predicted Tree}} &
        
        \includegraphics[width=.15\linewidth]{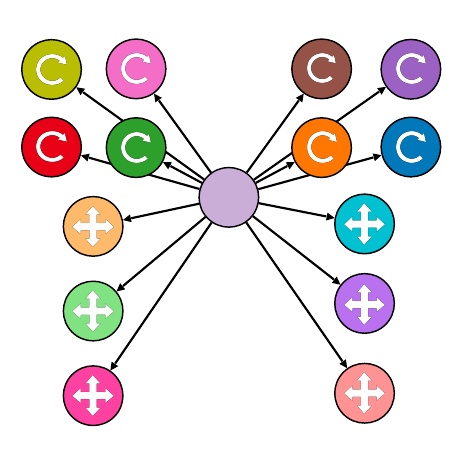} &
        \includegraphics[width=.15\linewidth]{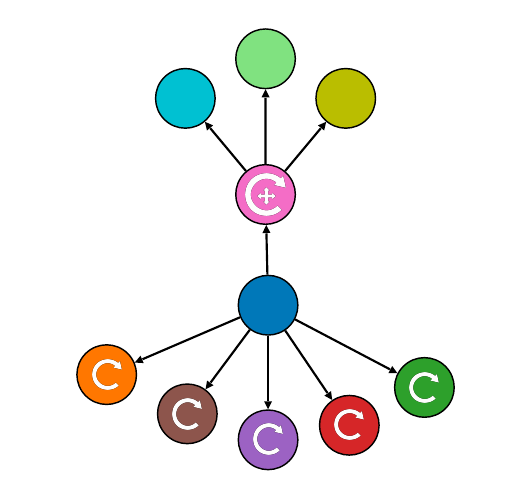} &
        \includegraphics[width=.15\linewidth]{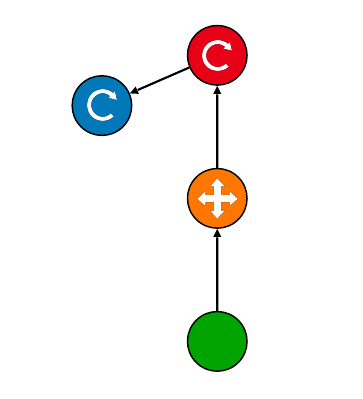} &
        \includegraphics[width=.15\linewidth]{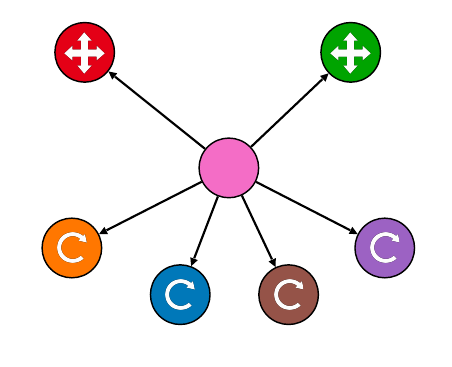} &
        \includegraphics[width=.15\linewidth]{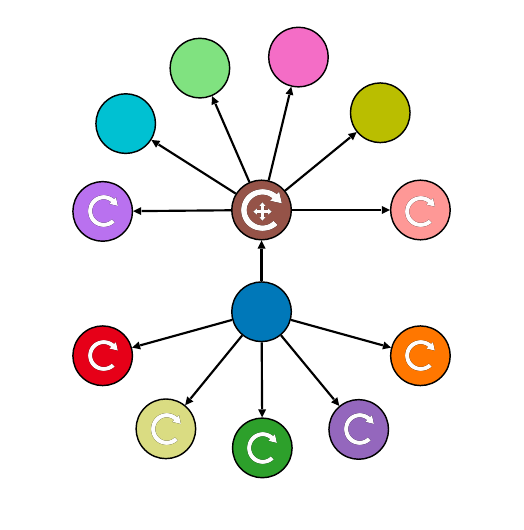} &
        \includegraphics[width=.15\linewidth]{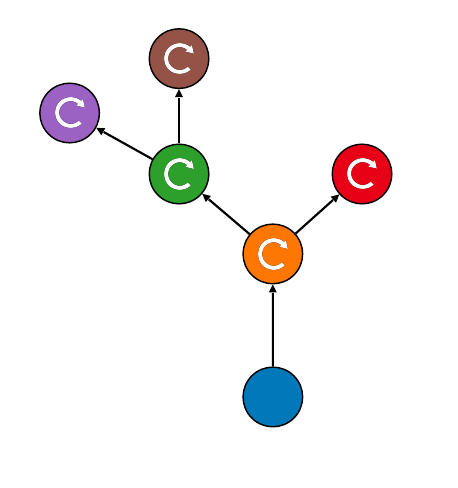}  
        \\
        \raisebox{1.5em}{\rotatebox{90}{Target Tree}} &
        \includegraphics[width=.15\linewidth]{figures/fig7/fig6-stg-clean.pdf} &
        \includegraphics[width=.15\linewidth]{figures/fig7/fig6-chair-clean.pdf} &
        \includegraphics[width=.15\linewidth]{figures/fig7/fig6-lamp-clean.pdf} &
        \includegraphics[width=.15\linewidth]{figures/fig7/fig6-stg-noisy.pdf} &
        \includegraphics[width=.15\linewidth]{figures/fig7/fig6-chair-noisy.pdf} &
        \includegraphics[width=.15\linewidth]{figures/fig7/fig6-lamp-noisy.pdf}  
    \end{tabular}
    \caption{
    Examples of predicted kinematic hierarchies for both Clean and Noisy point clouds.
    All examples shown use ground truth part segmentations.
    }
    \label{fig:hierarchy_results}
\end{figure*}

We next evaluate our graph neural network module's ability to infer the structure of an object's kinematic hierarchy.
To assess this performance, we use four metrics:
\begin{itemize}
    \item $\mathbf{E}_\text{type}$: \% of nodes whose motion type is mis-predicted.
    \item $\mathbf{E}_\text{exist}$: \& of edges whose existence is mis-predicted.
    \item $\mathbf{E}_\text{dir}$: \% of edges whose direction is mis-predicted.
    \item $\mathbf{E}_\text{root}$: \% of models whose root node is mis-predicted.
    \item \textbf{Tree F1}: the F1 score of the predicted kinematic hierarchy with respect to the ground truth one. Precision is computed via top-down traversal of the predicted tree, counting the fraction of nodes and edges which match their counterparts in the ground truth tree. Recall computation uses the same procedure, with the roles of the predicted and ground-truth trees reversed.
\end{itemize}
Table~\ref{tab:hierarchy_ablation} summarizes the performance of our network on Clean and Noisy synthetic data.
As expected, the GNN evaluated on clean data with ground truth instance segmentation preforms well on all classes. Even in the case where $\mathbf{E}_\text{dir}$ is relatively high for the lamp and chair class the negative log probability pairwise matrix complemented with MST is able to construct trees with a negligible amount of error. 
On noisy data, performance again degrades slightly. 
Fig.~\ref{fig:hierarchy_results} shows some qualitative examples. 


\begin{table}[ht]
    \centering
    \setlength{\tabcolsep}{1.25pt}
    \caption{Kinematic structure error for different data types}
    \begin{tabular}{lllcccc}
        \toprule
        \textbf{Category} & \textbf{Data Type} &  $\mathbf{E_\text{type}}$ & $\mathbf{E_\text{exist}}$ & $\mathbf{E_\text{dir}}$ & $\mathbf{E_\text{root}}$ & \textbf{Tree F1}
        \\
        \midrule
        \multirow{2}{*}{\emph{Storage Furniture}} & Clean &   1.16	& 1.2 &	2.22 &	0 &	99.73
        \\
        & Noisy &  0.39 & 2.03 & 29.04 & 0 & 99.02
        \\
        \midrule
        \multirow{2}{*}{\emph{Lamp}} & Clean &  0.47& 4.63& 31.11& 0.23& 98.24
        \\
        & Noisy &  0 & 0.98 &17.24	& 0.92	& 97.82
        \\
        \midrule
        \multirow{2}{*}{\emph{Chair}} & Clean &  0.08 & 0.1 & 39.64 & 0 & 99.77
        \\
        & Noisy &  0.24 & 0.99 & 41.73 & 0 & 98.67
        \\
        \bottomrule
    \end{tabular}
    \label{tab:hierarchy_ablation}
\end{table}

\begin{figure}[t!]
    \centering
    \vspace{0.25cm}
    \setlength{\tabcolsep}{1pt}
    \begin{tabular}{ccc}
        \multicolumn{3}{c}{\rule[1.5pt]{9em}{0.8pt} Scanning \rule[1.5pt]{9em}{0.8pt}}
        \\
        \includegraphics[width=.31\linewidth]{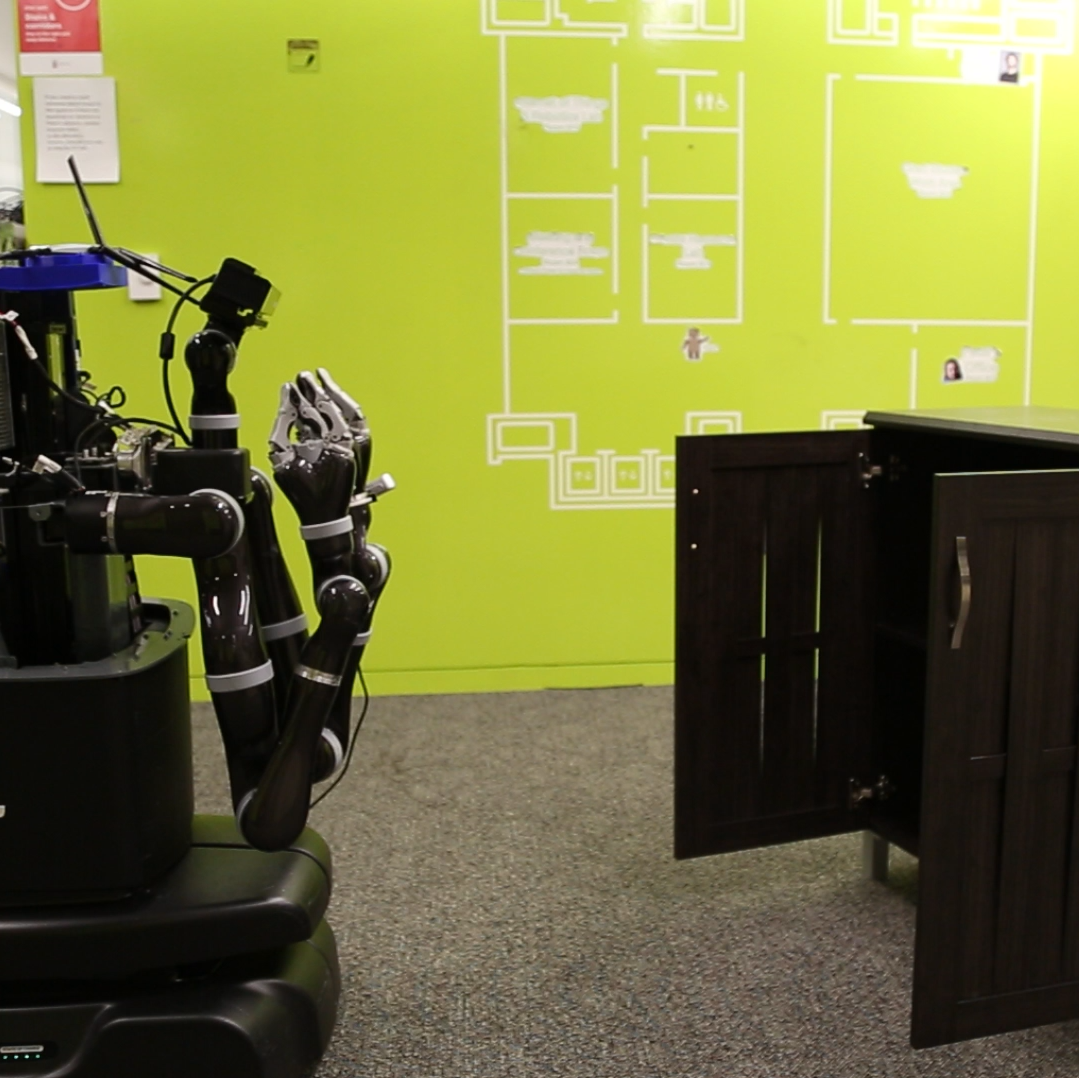} &
        \includegraphics[width=.31\linewidth]{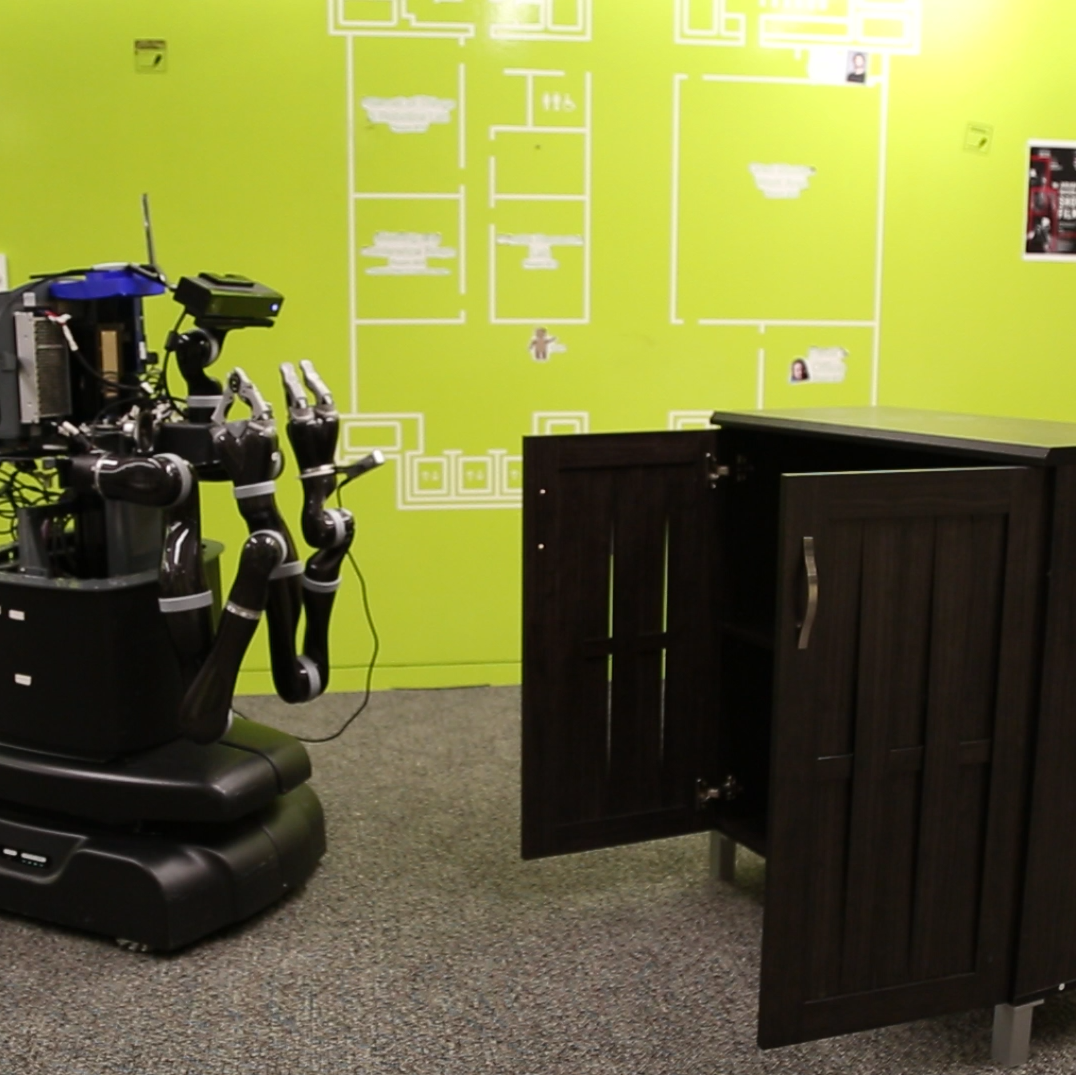} &
        \includegraphics[width=.31\linewidth]{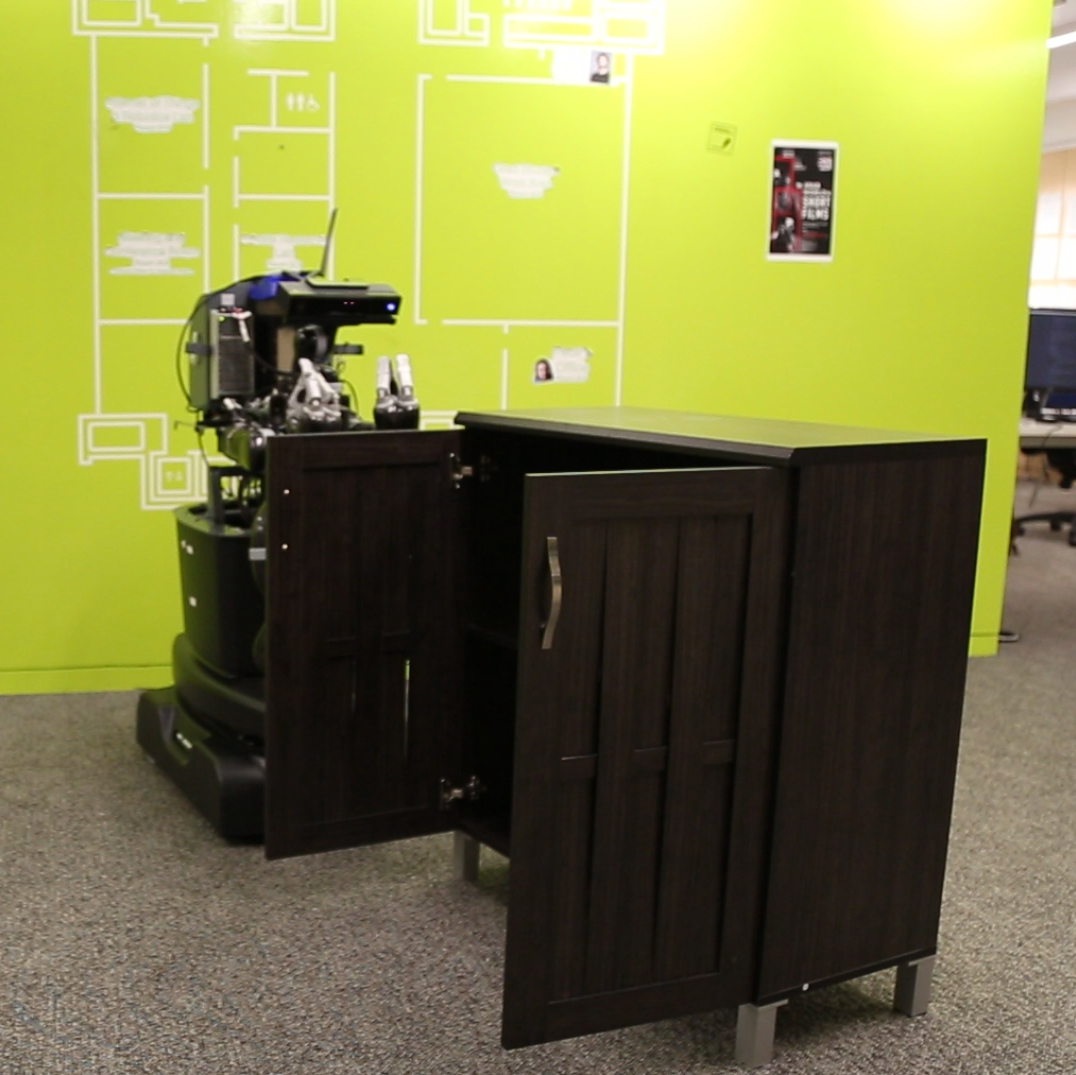}
        \\
        \multicolumn{3}{c}{\rule[1.5pt]{9em}{0.8pt} Interacting \rule[1.5pt]{9em}{0.8pt}}
        \\
        \includegraphics[width=.31\linewidth]{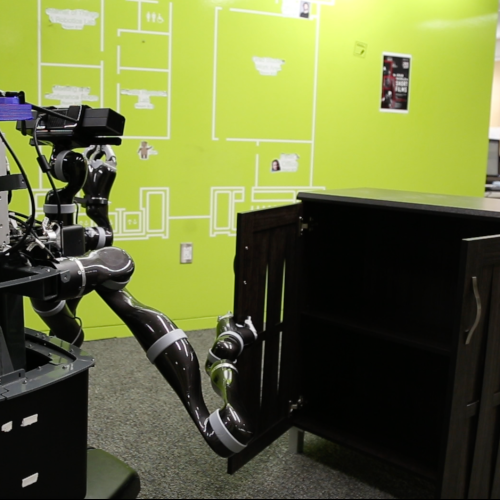} &
        \includegraphics[width=.31\linewidth]{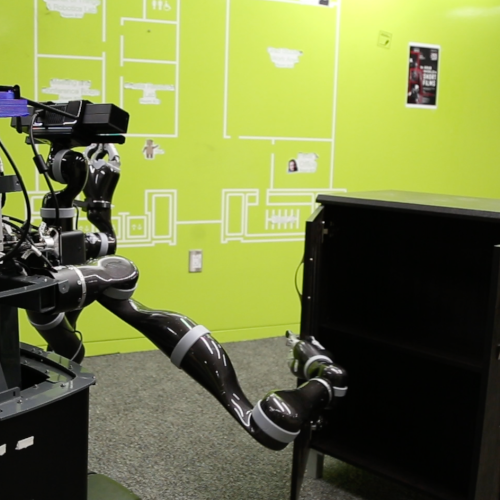} &
        \includegraphics[width=.31\linewidth]{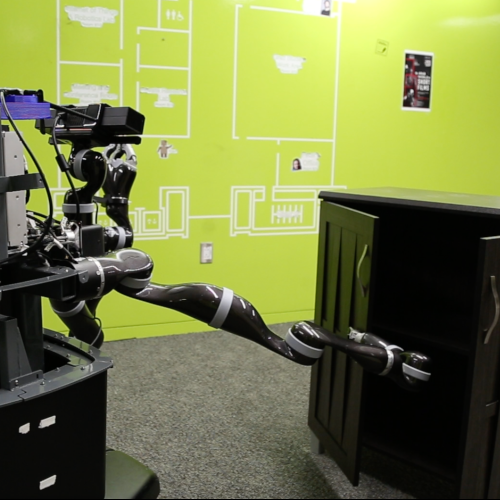}
    \end{tabular}
    \caption{Frames from a video demonstrating a robot scanning and then interacting with an articulated object.}
    \label{fig:robot_demo}
\end{figure}

\subsection{Robot Demo}
Finally, we demonstrate that the model enables a Kinova Movo robot to manipulate novel object instances. 
A storage furniture object was densely scanned using the robot's Kinect sensor by driving the robot around the object. Pointclouds were merged using RTAB-Map \cite{labbe2019rtab}, which performed SLAM using the robot's odometry, base laser scans, and visual data. 
The object was manually segmented from the scene and assigned a frame of reference. 
Our model successfully performed part segmentation and kinematic hierarchy inference; predictions are visualized in Figure~\ref{fig:teaser}. 
Motion model predictions, e.g. the axes of rotation, were generated using type predictions from the estimated kinematic tree and the intersections of part bounding boxes. 
Using the predicted tree and motion models, the agent is able to plan and execute interactions with the object. 
We fixed a grasp on the relevant object part, and ran a simple controller which moves the end-effector along the predicted direction of motion toward a goal configuration. 
Fig.~\ref{fig:robot_demo} shows frames from a video of the robot scanning the cabinet and closing one of its doors.


%% file: sections/06-future.tex


\section{Future Work}
\label{sec:future}


Training our neural networks requires collections of part segmented, kinematically-annotated 3D models.
Such data is not widely available, as it requires nontrivial human annotation effort.
Recent computer vision research has demonstrated the possibility of segmenting 3D models with limited or even no human supervision~\cite{BAENet,AdaCoSeg}.
The development of similar techniques for identifying articulated motion within 3D shapes would allow our methods to be applied to any object type for which unsegmented 3D models are available~\cite{xu2020motion}.

Our approach treats segmentation and graph labeling as separate sub-problems, but one can argue that they are coupled: parts determines what motions are possible, and whether a motion is possible determines whether a proposed part is a good one.
Accordingly, the Shape2Motion system found benefits in jointly segments and predict motion parameters for parts~\cite{Shape2Motion}.
It is possible that jointly segmenting and predicting kinematic hierarchy could confer similar benefits.

We focused on inferring the structure of an object's kinematic hierarchy.
Our system could be combined with one that focuses on kinematic motion parameter prediction~\cite{Shape2Motion,RPMNet,DeepPart} to produce a complete mobility perception system.



%% file: sections/07-conclusion.tex


\section{Conclusion}
\label{sec:conclusion}


We presented a perception system that infers kinematic hierarchies for never-before-seen object instances.
Our system infers the moving parts of an object
and the kinematic couplings between them.
To infer parts, it uses a point cloud segmentation neural network, improved via a multi-view consensus algorithm.
To infer kinematic couplings, it uses a graph neural network to predict the existence, direction, and type of edges (i.e. joints) that relate the inferred parts.
We train these networks using simulated scans of synthetic 3D models.
In experiments, our system inferred accurate kinematic hierarchies for simulated scans of 3D objects and scans of real-world objects gathered by a mobile robot.
